\documentclass[journal,transmag]{IEEEtran}
\usepackage{cite}
\ifCLASSINFOpdf
  \usepackage[pdftex]{graphicx}
\else
\fi
\usepackage{amsmath}
\usepackage{algorithmic}
\usepackage{array}
\ifCLASSOPTIONcompsoc
 \usepackage[caption=false,font=normalsize,labelfont=sf,textfont=sf]{subfig}
\else
 \usepackage[caption=false,font=footnotesize]{subfig}
\fi
\usepackage{fixltx2e}
\usepackage{dblfloatfix}
\ifCLASSOPTIONcaptionsoff
 \usepackage[nomarkers]{endfloat}
\let\MYoriglatexcaption\caption
\renewcommand{\caption}[2][\relax]{\MYoriglatexcaption[#2]{#2}}
\fi
\usepackage{url}
\newcommand\Tstrut{\rule{0pt}{2.0ex}}         % = `top' strut
\newcommand\Bstrut{\rule[-0.9ex]{0pt}{0pt}}   % = `bottom' strut

\usepackage{color}
\usepackage[ruled,vlined,algo2e]{algorithm2e}
\SetKwComment{Comment}{$\triangleright$\ }{}

\usepackage{pifont}
\newcommand{\cmark}{\ding{51}}%
\newcommand{\xmark}{\ding{55}}%

\usepackage{multirow}
\usepackage{hyperref}

\begin{document}
\title{POUR-Net: A Population-Prior-Aided Over-Under-Representation Network for Low-Count PET Attenuation Map Generation}


\author{Bo Zhou, Jun Hou, Tianqi Chen, Yinchi Zhou, Xiongchao Chen, Huidong Xie, Qiong Liu, Xueqi Guo, \\ 
Yu-Jung Tsai, Vladimir Y. Panin, Takuya Toyonaga, James S. Duncan, Chi Liu
\thanks{B. Zhou and X. Chen and H. Xie and Q. Liu and X. Guo are with the Department of Biomedical Engineering, Yale University, New Haven, CT, 06511, USA. J. Hou and T. Chen are with the Department of Radiology and Biomedical Imaging, Yale University, New Haven, CT, 06511, USA, and the Department of Computer Science, University of California Irvine, Irvine, CA, 92697, USA. Y. Tasi was with the Department of Radiology and Biomedical Imaging, Yale University, New Haven, CT, 06511, USA, and is currently with Canon Medical Research USA Inc., Vernon Hills, IL, 60061, USA. V. Y. Panin is with the Siemens Medical Solutions Inc., Knoxville, TN, 37932, USA. T. Toyonaga is with the Department of Radiology and Biomedical Imaging, Yale University, New Haven, CT, 06511, USA. J. S. Duncan and C. Liu are with the Department of Biomedical Engineering and the Department of Radiology and Biomedical Imaging, Yale University, New Haven, CT, 06511, USA. 
Corresponding email: bo.zhou@yale.edu and chi.liu@yale.edu}
\thanks{This work was supported by funding from the National Institutes of Health (NIH) under grant numbers R01EB025468, R01CA224140, and R01CA275188. }
}

% \markboth{Journal of \LaTeX\ }%



% for Transactions on Magnetics papers, we must declare the abstract and
% index terms PRIOR to the title within the \IEEEtitleabstractindextext
% IEEEtran command as these need to go into the title area created by
\IEEEtitleabstractindextext{%
\begin{abstract}
Low-dose PET offers a valuable means of minimizing radiation exposure in PET imaging. However, the prevalent practice of employing additional CT scans for generating attenuation maps ($\mu$-map) for PET attenuation correction significantly elevates radiation doses. To address this concern and further mitigate radiation exposure in low-dose PET exams, we propose POUR-Net - an innovative population-prior-aided over-under-representation network that aims for high-quality attenuation map generation from low-dose PET. First, POUR-Net incorporates an over-under-representation network (OUR-Net) to facilitate efficient feature extraction, encompassing both low-resolution abstracted and fine-detail features, for assisting deep generation on the full-resolution level. Second, complementing OUR-Net, a population prior generation machine (PPGM) utilizing a comprehensive CT-derived $\mu$-map dataset, provides additional prior information to aid OUR-Net generation. The integration of OUR-Net and PPGM within a cascade framework enables iterative refinement of $\mu$-map generation, resulting in the production of high-quality $\mu$-maps. Experimental results underscore the effectiveness of POUR-Net, showing it as a promising solution for accurate CT-free low-count PET attenuation correction, which also surpasses the performance of previous baseline methods.
\end{abstract}

\begin{IEEEkeywords}
Low-count PET, Attenuation Correction, Population Prior, Deep Learning, Attenuation Map Generation
\end{IEEEkeywords}}

\maketitle

\IEEEdisplaynontitleabstractindextext

% \fi
\IEEEpeerreviewmaketitle

\section{Introduction}
% Intro to PET and the importance of low-count PET
% Further lowering the dose with CT-less attenuation map generation & other benefits
% Previous works in u-map generation in PET (MLAA - DL)
% No work on low-count PET attenuation map generation - directly using prior method may not sufficient
% What we done (UO-Net) for enhance fine detail learning & population prior to aid the process
% What we found ....

\IEEEPARstart{P}{ositron} emission tomography (PET) is an important functional imaging modality with wide clinical applications in oncology, cardiology, and neurology. To perform PET imaging, a small amount of radioactive tracer is injected into the patient, which inevitably introduces radiation exposure to both patient and healthcare providers, thus raising concerns. By reducing the administered injection dose, low-dose/count PET (referred to as low-count PET hereafter) is of great interest per the As Low As Reasonably, Achievable concept (ALARA) \cite{strauss2006alara}, in particular for clinical applications that require serial PET scanning, e.g. therapy response evaluation \cite{ben200918f}, or patient population with limited tolerance to radiation, e.g. pediatric PET \cite{parisi2017optimization}. Even though low-count PET with conventional reconstruction methods, e.g. OSEM \cite{lantos2018standard}, suffer from low image quality, recent developments of statistical-based \cite{dutta2013non,maggioni2012nonlocal,mejia2016noise} and deep learning (DL)-based PET denoising/reconstruction methods \cite{wang20183d,kaplan2019full,hu2020dpir,gong2020parameter,zhou2020supervised,chen2019ultra,zhou2021mdpet,xie2023unified,zhou2022federated,zhou2023fedftn,xie2023dose} have shown great potential to reconstruct high-quality PET image from low-count data, and could provide highly consistent quantification when comparing to standard-count PET. On the other hand, accurate PET quantification also relies on accurate Attenuation Correction (AC) during PET reconstruction. In PET/CT scans, the CT imaging component aims to provide a high-quality attenuation map, i.e. $\mu$-map, for AC. However, this AC protocol not only introduces additional radiation for PET imaging procedures but may cause various artifacts in the reconstructed PET. For example, these artifacts can originate from PET-CT misalignments \cite{martinez2007artifacts,lu2018respiratory}, or CT itself \cite{boas2012ct}, e.g. due to beam hardening, metal artifact, and count-starving, and low-dose CT may exacerbate those artifacts \cite{zhou2021dudodr,zhou2022dudoufnet,zhou2020limited}. To further reduce the radiation in PET scans and improve the PET image quality, it is desirable to develop CT-free AC methods and generate a high-quality $\mu$-map directly from low-count PET for AC. 

% \cite{xiang2017deep,wang20183d,kaplan2019full,hu2020dpir,gong2020parameter,zhou2020supervised,ouyang2019ultra,chen2019ultra,liu2020noise,zhou2021mdpet,xie2023unified,zhou2022federated,zhou2023fedftn,xie2023dose}

In fact, for time-of-flight (TOF) PET, Maximum Likelihood Reconstruction of Activity and Attenuation (MLAA) algorithms \cite{rezaei2016simultaneous} were first developed to simultaneously reconstruct the tracer activity ($\lambda$-MLAA) and the attenuation map ($\mu$-MLAA), based on the TOF PET raw data only, i.e. without CT. Ideally, MLAA would solve the AC problem for TOF PET. However, the direct use of $\mu$-MLAA still suffers from significant quantification error as compared to the CT-based ($\mu$-CT) OSEM reconstruction \cite{rezaei2016simultaneous,shi2023deep,nuyts2018validation}, even in the standard-count PET. With recent advancements in deep learning and computer vision, DL-based approaches have been extensively studied to address this challenge \cite{lee2020review,chen2023deep}. The previous DL-based AC methods can be summarized into three classes. The first category of methods uses either non-attenuation corrected (NAC) PET or MLAA reconstructions as deep network input to generate pseudo-CT or CT-based $\mu$-map \cite{shi2023deep,toyonaga2022deep,liu2018deep1,liu2018deep2,hwang2019generation,dong2019synthetic}. Then, AC is performed based on the generated $\mu$-map. Similarly, the second category of methods also uses either NAC PET or MLAA reconstructions as deep network input, but to directly generate attenuation-corrected PET \cite{shiri2019direct,dong2020deep,shiri2023differential}. The above two classes of methods are easy to implement and can be applied to all PET modalities, i.e. PET-alone, PET/CT, or PET/MRI, which however suffer from quantification errors given the relatively low amount of information if input is from the NAC PET only. The third category of methods was mainly for PET/MRI, where deep networks were developed for synthesizing pseudo-CT or CT-based $\mu$-map based on the MRI images \cite{bradshaw2018feasibility,leynes2018zero,arabi2019novel,spuhler2019synthesis}. However, these methods are subject to MRI artifacts and PET/MRI misalignment errors \cite{zhuo2006mr}. Most importantly, almost all these previous methods were developed based on U-Net or its variants\cite{ronneberger2015u,isola2017image,zhu2017unpaired} and only validated on standard-count PET. While these methods achieved reasonable AC performance on standard-count PET, directly deploying these methods for low-count PET AC may result in sub-optimal performance, given 1) the low-quality inputs of either NAC PET or MLAA, and 2) insufficient feature representation learned by encoder-decoder network and its variants, i.e. U-Net, from the low-count input(s). Therefore, a network with rich feature representation is essential for high-quality AC in low-count PET. 

In this work, we focused on high-quality $\mu$-map generation from low-count PET data, and then utilized this virtual $\mu$-map for low-count PET AC. The motivation is two-fold. First, previous works for AC in PET and SPECT found superior performance from the indirect AC approach, i.e. using the generated $\mu$-map for AC, better than the direct AC approach, i.e. directly generating the attenuation corrected PET. Second, after generating the low-count PET with indirect AC, previous low-count PET denoising methods are directly applicable to reconstructing low-count PET to stand-count PET images.

\begin{figure*}[htb!]
\centering
\includegraphics[width=0.92\textwidth]{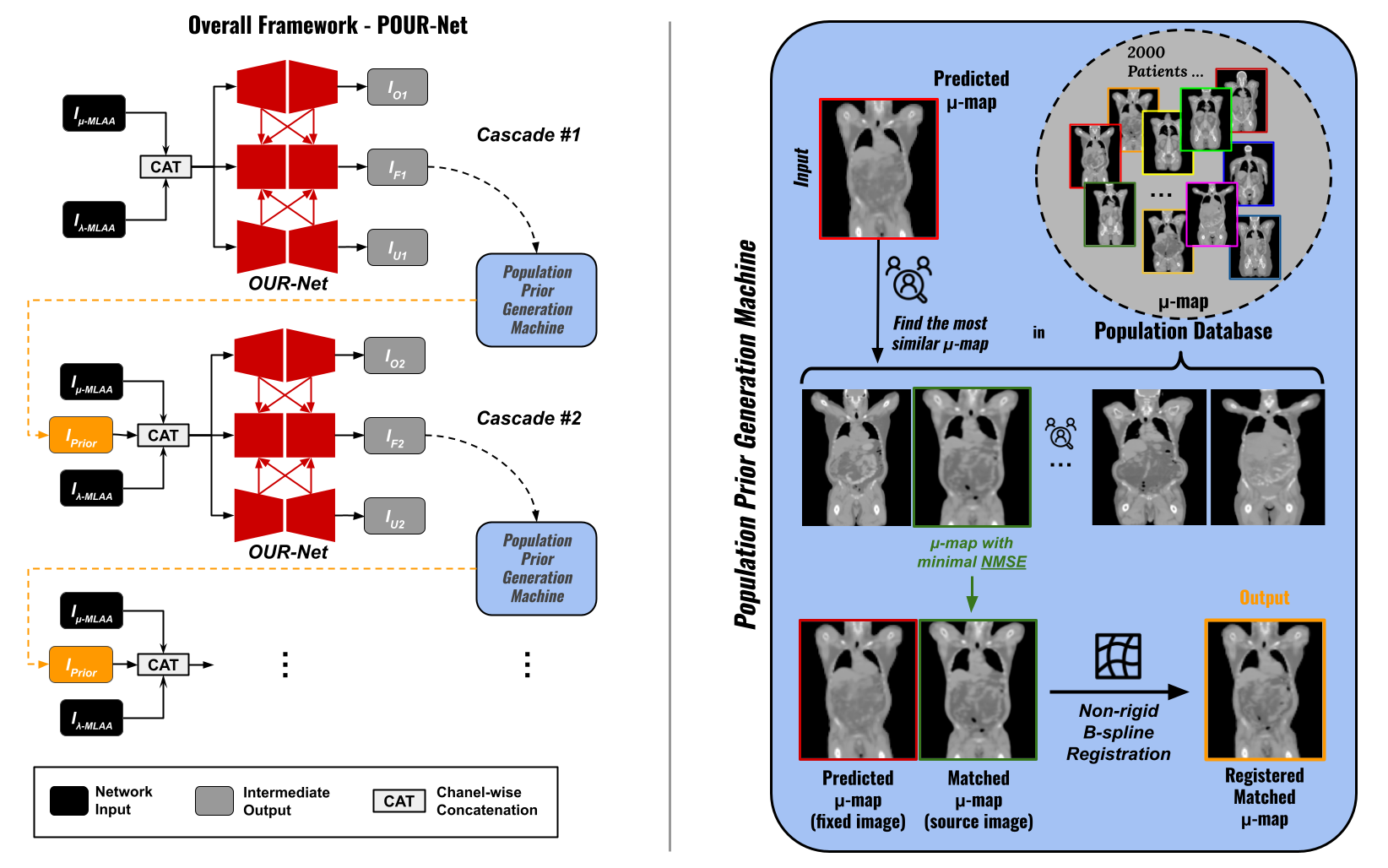}
\caption{The overall pipeline of our population prior-aided over-under-representation network (POUR-Net) for attenuation map ($\mu$-map) generation. POUR-Net integrates the over-under-representation network (OUR-Net) and Population-Prior Generation Machine (PPGM) for iterative refinement of $\mu$-map generation (left part). The detailed steps of PPGM are illustrated on the right.}
\label{fig:framework}
\end{figure*}

To generate high-quality $\mu$-maps from low-count PET, we proposed a population-prior-aided over-under-representation network, called POUR-Net. The main contribution is three-fold. First, to enable efficient feature extraction and representation learning from low-count PET inputs, we proposed an over-under-representation network (OUR-Net) that can extract both low-resolution abstracted features from the under-representation branch and fine-detail features from the over-representation branch for assisting the $\mu$-map generation. Second, given widely available CT scans from different patient exams with or without PET, we collected a large-scale CT-derived $\mu$-map dataset and designed a population prior generation machine (PPGM). The output of this can be used as input of the network to aid the DL-based $\mu$-map generation. Lastly, we integrated the OUR-Net and PPGM into a cascade framework that allows iterative refinement of $\mu$-map generation (Figure \ref{fig:framework}). Our experimental results show that POUR-Net can generate high-quality $\mu$-map under different low-count/dose PET settings. Using the generated $\mu$-map, we also found that our method can generate highly consistent attenuation-corrected PET as compared to those with CT-based AC, better than previous DL baselines. 

%===========================================================
\section{Methods}
% \subsection{Overview} 
The overall framework of POUR-Net is depicted in Figure \ref{fig:framework}. POUR-Net aims to generate high-quality $\mu$-map from low-count PET data using a cascade design consisting of two key components, including an Over-Under-Representation Network (OUR-Net) and a Population-Prior Generation Machine (PPGM). 

Specifically, the framework inputs are $\lambda$-MLAA and $\mu$-MLAA reconstructed from MLAA \cite{rezaei2016simultaneous} which contain coarse estimations of tracer activity information and attenuation information, respectively. In the initial cascade, the input to the OUR-Net is the channel-wise concatenated $\lambda$-MLAA and $\mu$-MLAA. While a reasonable $\mu$-map could be generated, it is still difficult for the network to recover the fine details, e.g. rib and spine structures, directly from the low-count MLAA inputs, given the noisy inputs with artifacts. Thus, the predicted $\mu$-map from the first cascade is then input into the PPGM, along with a large-scale population dataset of CT-derived $\mu$-map that is independent of the PET patient cohort, to retrieve additional information for aiding the $\mu$-map generation. In PPGM, we search for the most matched $\mu$-map from this independent cohort and then register it to the input of network prediction. The resulting registered $\mu$-map retrieved from the stand-alone dataset serves as the population-prior $\mu$-map, functioning as an additional input to the OUR-Net in the next cascade. Throughout this iterative process, the $\mu$-map generation can be gradually refined with the assistance of the PPGM. Each instance of OUR-Net in this cascaded framework is trained independently. Details about OUR-Net and PPGM are described in detail in the subsequent sections. 

\subsection{Over-Under-Representation Network} 
\begin{figure*}[htb!]
\centering
\includegraphics[width=0.96\textwidth]{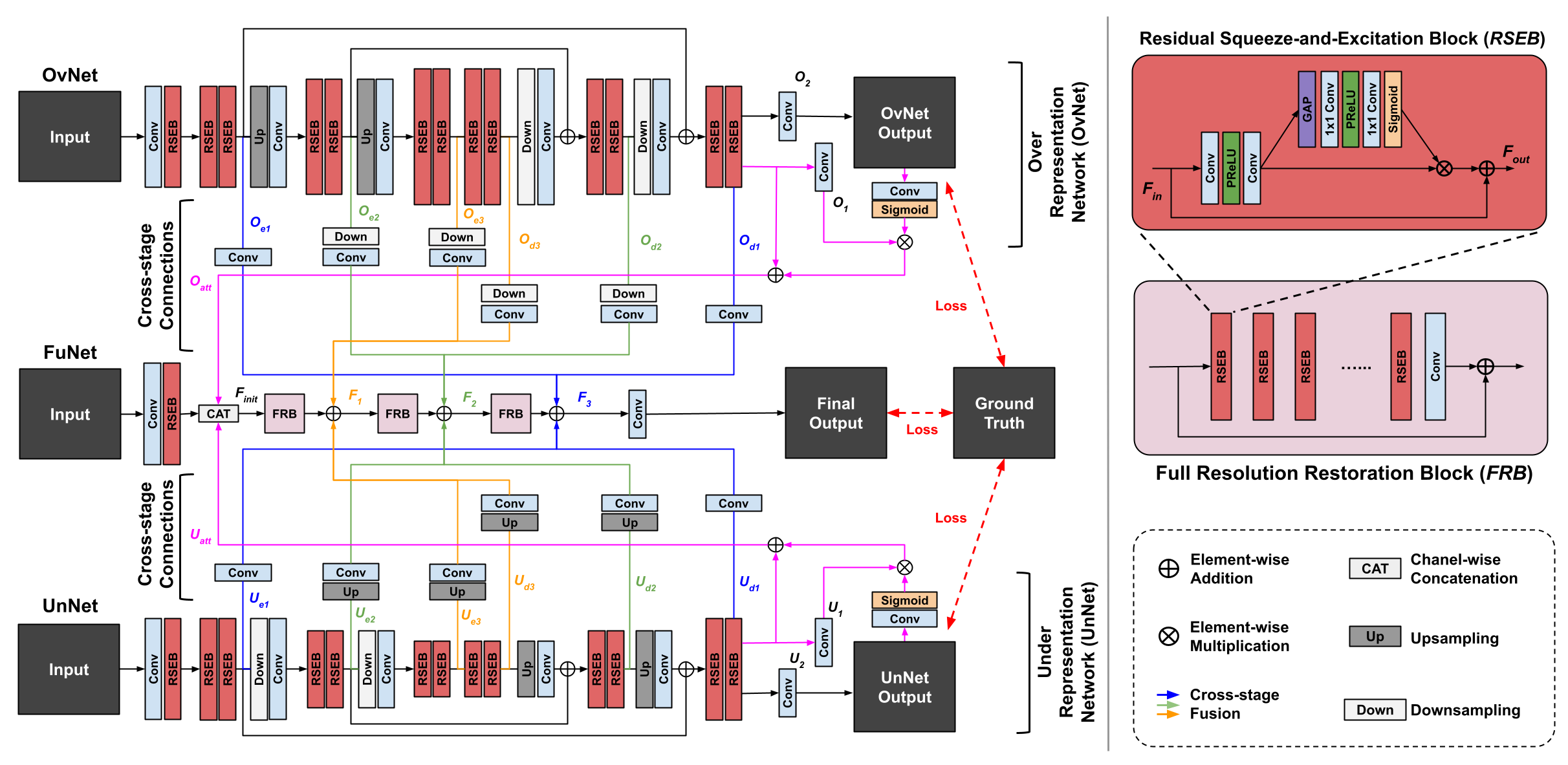}
\caption{The detailed network architecture of the over-under-representation network (OUR-Net) used in the cascade framework of POUR-Net (Figure \ref{fig:framework}). The network consists of an over-represented network (OvNet) branch at the top and an under-represented network (UnNet) branch at the bottom, for assisting the generation in the central full-resolution network (FuNet) branch.}
\label{fig:network}
\end{figure*}

The main network used in our cascaded framework is the Over-Under-Representation Network (OUR-Net). OUR-Net uses a three-stage network architecture, consisting of an under-representation subnetwork branch (UnNet) and an over-representation sub-network branch (OvNet). The UnNet focuses on abstracted feature extractions, while the OvNet focuses on over-representing fine-detail extractions. The combined features of UnNet and OvNet are inputted into the fully-representation subnetwork (FuNet) for final output. The architecture of OUR-Net is illustrated in Figure \ref{fig:network}. 

Specifically, UnNet is based on a U-shape network \cite{ronneberger2015u} for under-representation feature extraction, as illustrated in the bottom part of Figure \ref{fig:network}. On the other hand, OvNet is also based on a U-shape network, but with a reverse order of downsampling and upsampling operations, as illustrated in the upper part of Figure \ref{fig:network}. For feature extraction at each resolution level in both UnNet and OvNet, we use a residual squeeze-and-excitation block (RSEB) for feature extraction (red block in Figure \ref{fig:network}). Given an input feature $F_{in}$ for RSEB, the output can be written as:
\begin{equation}
    F_{out} = F_{in} + P_{se} [ P_{ex} (F_{in}) ] ,
\end{equation}
where $P_{ex}$ consists of two convolutional layers for feature extraction, and $P_{se}$ is the squeeze-and-excitation layer \cite{hu2018squeeze}, generating channel-attention from the input feature for channel-wise feature re-calibration. Given an input of $X_{in}$ for UnNet and OvNet, we first use a convolutional layer and an RSEB for initial feature extractions. Then, two consecutive RSEBs are employed for feature extraction at 3 different levels of UnNet and OvNet, generating 1) UnNet's encoding features of $U_{e1}$, $U_{e2}$, $U_{e3}$ and decoding features of $U_{d3}$, $U_{d2}$, $U_{d1}$, and 2) OvNet's encoding features of $O_{e1}$, $O_{e2}$, $O_{e3}$ and decoding features of $O_{d3}$, $O_{d2}$, $O_{d1}$. The multi-scale abstracted features from UnNet are helpful for general appearance reconstruction, while the multi-scale fine resolution features from OvNet are helpful for fine detail reconstruction. The final output of UnNet and OvNet then can be written as:
\begin{equation}
    X_{U} = P_{U} (U_{d1}) , 
\end{equation}
\begin{equation}
    X_{O} = P_{O} (O_{d1}) , 
\end{equation}
where $P_{U}$ and $P_{O}$ are both one level of convolutional layer for reducing the number of feature channels to 1. 

To use the final supervised features of UnNet and OvNet for the $\mu$-map generation at the central branch, we use a self-attention connection mechanism, i.e. the pink connections in Figure \ref{fig:network}, which can be formulated as:
\begin{equation}
    U_{att} = U_{d1} + P_{U_1} (U_{d1}) \odot \sigma (P_{U_2} (X_U)) ,
\end{equation}
\begin{equation}
    O_{att} = O_{d1} + P_{O_1} (O_{d1}) \odot \sigma (P_{O_2} (X_O)) ,
\end{equation}
where $P_{U_1}$ and $P_{O_1}$ are both a $3 \times 3$ convolutional layer for generating the unweighted features for UnNet and OvNet. $P_{U_2}$ and $P_{O_2}$ are also both a $3 \times 3$ convolutional layer for generating spatial-wise attention weights with inputs of $X_U$ and $X_O$. The attention weight is normalized by the sigmoid function $\sigma$, such that the attention weights lie between 0 and 1. 

Given the self-attention features of $U_{att}$ and $O_{att}$ and the same initial input $X_{in}$, an initial feature in the central branch of FuNet can be constructed as:
\begin{equation}
    F_{init} = \{ U_{att}, O_{att}, P_{init}(X_{in}) \} ,
\end{equation}
where $\{\}$ denotes channel-wise concatenation operation, and $P_{init}$ consists of a convolutional layer followed by an RSEB. $F_{init}$ is then input into three consecutive full-resolution restoration blocks (FRB). The intermediate features of FRBs fuse with the multi-scale features from UnNet and OvNet, so that the abstracted and fine detail features of UnNet and OvNet are utilized to assist the $\mu$-map generation on the original resolution. The process can be written as:
\begin{align}
    F_{1} = & P_{F1} (F_{init}) + P_{O_{e3}} (O_{e3}) \nonumber \\
            &  + P_{O_{d3}} (O_{d3}) + P_{U_{e3}} (U_{e3}) + P_{U_{d3}} (U_{d3}) , \\
    F_{2} = & P_{F2} (F_{1}) + P_{O_{e2}} (O_{e2}) \nonumber \\
            & + P_{O_{d2}} (O_{d2}) + P_{U_{e2}} (U_{e2}) + P_{U_{d2}} (U_{d2}) , \\
    F_{3} = & P_{F3} (F_{2}) + P_{O_{e1}} (O_{e1}) \nonumber \\
            &  + P_{O_{d1}} (O_{d1}) + P_{U_{e1}} (U_{e1}) + P_{U_{d1}} (U_{d1})
\end{align}
where $P_{F1}$, $P_{F2}$, and $P_{F3}$ are FRB, consisting of multiple RSEB with residual connections between the block's input and output (pink block in Figure \ref{fig:network}). For UnNet's intermediate features, $P_{U_{e3}}$ and $P_{U_{d3}}$ consists of a $\times 4$ upsampling operation followed by a convolutional layer (yellow lines extended from UnNet in Figure \ref{fig:network}), while $P_{U_{e2}}$ and $P_{U_{d2}}$ consists of a $\times 2$ upsampling operation followed by a convolutional layer (green lines extended from UnNet). $P_{U_{e1}}$ and $P_{U_{d1}}$ only contain one convolutional layer (blues lines extended from UnNet). Similarly, for OvNet's intermediate features, $P_{O_{e3}}$ and $P_{O_{d3}}$ consists of a $\times 4$ downsampling operation followed by a convolutional layer (yellow lines extended from OvNet in Figure \ref{fig:network}), while $P_{O_{e2}}$ and $P_{O_{d2}}$ consists of a $\times 2$ downsampling operation followed by a convolutional layer (green lines extended from OvNet). $P_{O_{e1}}$ and $P_{O_{d1}}$ only contain one convolutional layer (blue lines extended from OvNet). Finally, the output of FuNet can be written as $X_F = P_{F} (F_{3})$, where $P_{F}$ is a single convolutional layer for reducing the number of feature channels to 1. Here, we empirically set the number of RSEB to 4 in FRB, such that FuNet contains approximately the same amount of parameters as UnNet and OvNet, as well as for efficient 3D computing. 

Since the FuNet operates on the original resolution, it can maintain high-resolution features. The multi-scale feature from UnNet and OvNet helps enrich the feature here, thus aiding high-quality $\mu$-map generation. For network training, supervision is applied to all subnetwork's outputs. The objective function thus can be written as:
\begin{equation}
    \mathcal{L}_{tot} = || X_F - X_{gt} ||_2^2 + || X_{U} - X_{gt} ||_2^2  + || X_{O} - X_{gt} ||_2^2 ,
\end{equation}
where $X_{gt}$ is the CT-derived $\mu$-map. From the OUR-Net, $X_F$ is the final network output. 

\subsection{Population-Prior Generation Process}
Since it is challenging to directly generate $\mu$-map from the noisy and artifact-present low-count PET MLAA images, we further harness the information available from a large population of CT data, acquired with or without PET scan, for assisting OUR-Net's $\mu$-map generation. The population prior generation process is summarized in the blue block of Figure \ref{fig:framework}. Specifically, we assume a large-scale CT-derived $\mu$-map dataset of $\mathcal{D} = \{I_1, I_2, \cdots, I_N\}$, where $N$ is the number of patient here. Given an initial $\mu$-map produced by OUR-Net, i.e. $X_F$, our PPGM undergoes a dual-stage refinement process. 

In the first stage, we search over $\mathcal{D}$ and identify the most closely matched CT-derived $\mu$-map from the patient cohort by minimizing MSE with the OUR-Net-predicted $\mu$-map:
\begin{equation}
     j = \arg \min_{i} ||X_F - I_i||_2^2
\end{equation}
where $I_i$ represents the CT-derived $\mu$-map of the $i$-th patient in the extensive dataset $\mathcal{D}$. In the second stage, non-rigid registration based on demons \cite{vercauteren2009diffeomorphic} is employed to align the matched $\mu$-map ($I_j$) with network predicted $\mu$-map:
\begin{equation}
    I_{Prior} = T_{Prior}(I_j)
\end{equation}
where the non-rigid transformation $T_{Prior}$ is computed by minimizing the dissimilarity metrics between the matched $\mu$-map and the network predicted $\mu$-map:
\begin{equation}
    T_{Prior} = \arg \min_{T} ||X_F - T \circ I_j||_2^2 + \mathcal{R}(T)
\end{equation}
where $\circ$ is the transformation operator, and $\mathcal{R}$ is the regulation term for constraining the deformation field. Finally, the PPGM output ($I_{Prior}$) is used as an addition channel input to the OUR-Net of the next stage for the generation of the $\mu$ map in the next cascade, as shown in the left of Figure \ref{fig:framework}. 

\begin{figure*}[htb!]
\centering
\includegraphics[width=0.90\textwidth]{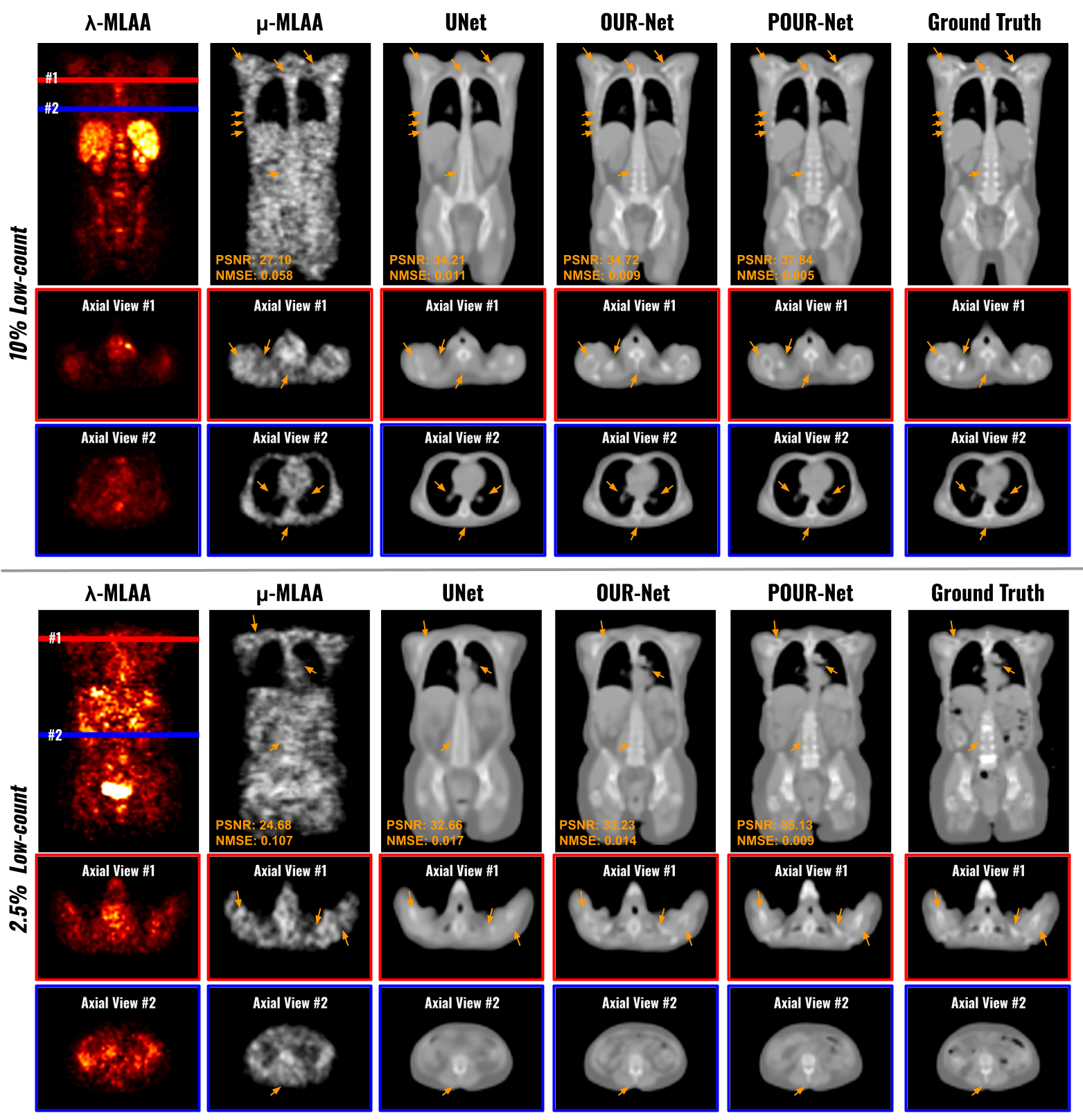}
\caption{Visual comparison of $\mu$-map generation from different methods under 10\% and 2.5\% low-count PET settings. The coronal view and selected axial views (red and blue cuts) are shown. RMSE and PSNR values are calculated for each individual volume, with the CT-derived $\mu$ map as a reference (last column).}
\label{fig:comp_methods_umap}
\end{figure*}

\begin{table*} [htb!]
\footnotesize
\centering
\caption{Quantitative comparison of $\mu$-map generation from different methods under two different low-count settings using PSNR, SSIM, and RMSE. Best results are marked in \textbf{bold}. "$\dagger$" means the differences between POUR-Net and all the baseline methods are significant at $p<0.001$.}
\label{tab:comp_methods_umap}
    \begin{tabular}{l|c|c|c||c|c|c}
        \hline
        \multirow{2}{*}{\textbf{Evaluation}}  & \multicolumn{3}{c}{\textbf{10\% Low-count}}                   &  \multicolumn{3}{c}{\textbf{2.5\% Low-count}}                    \Tstrut\Bstrut\\
        \cline{2-7}
                                              & PSNR             & SSIM              & RMSE              & PSNR            & SSIM              & RMSE                \Tstrut\Bstrut\\
        \hline
        $\mu$-MLAA \cite{rezaei2012simultaneous}                    & $26.82\pm1.23$   & $.899\pm.013$     & $.053\pm.022$     & $22.51\pm1.54$   & $.878\pm.021$  & $.143\pm.045$      \Tstrut\Bstrut\\
        \hline
        UNet\cite{toyonaga2022deep}                & $33.75\pm1.80$   & $.968\pm.006$     & $.012\pm.009$     & $31.19\pm2.07$   & $.959\pm.011$  & $.023\pm.013$       \Tstrut\Bstrut\\
        \hline
        RDUNet\cite{chen2021ct}              & $33.79\pm1.82$   & $.969\pm.007$     & $.012\pm.010$     & $31.27\pm2.05$   & $.960\pm.011$  & $.022\pm.013$        \Tstrut\Bstrut\\
        \hline
        cGAN\cite{armanious2020independent}            & $33.69\pm1.89$   & $.962\pm.008$     & $.013\pm.009$     & $31.11\pm2.10$   & $.958\pm.012$  & $.024\pm.014$       \Tstrut\Bstrut\\
        \hline
        MPRNet\cite{zamir2021multi}               & $33.92\pm1.88$   & $.970\pm.007$     & $.011\pm.009$     & $32.02\pm2.09$   & $.962\pm.011$  & $.021\pm.013$       \Tstrut\Bstrut\\
        \hline
        \hline
        OUR-Net                                 & $34.13\pm1.92$   & $.979\pm.006$    & $.010\pm.008$      & $32.69\pm2.09$   & $.968\pm.011$  & $.019\pm.012$     \Tstrut\Bstrut\\
        \hline
        POUR-Net                            & $\mathbf{37.01\pm1.82}$$^\dagger$   & $\mathbf{.981\pm.003}$$^\dagger$    & $\mathbf{.006\pm.003}$$^\dagger$    & $\mathbf{35.12\pm2.33}$$^\dagger$   & $\mathbf{.978\pm.006}$$^\dagger$    & $\mathbf{.010\pm.008}$$^\dagger$      \Tstrut\Bstrut\\
        \hline
    \end{tabular}
\end{table*}

\subsection{Data Preparation and Implementation Details}
We collected clinical patient data from the Yale New Haven Hospital (YNHH) for training and evaluation of our method. Specifically, we collected PET/CT patient data between July 2018 and February 2020. The data was acquired using a Siemens Biograph mCT scanner at YNHH. PET scans were performed approximately 60 mins after intravenous injection of about 10 mCi \textsuperscript{18}F-FDG tracer, with 1-pass whole-body continuous bed scanning protocol. Based on careful examinations by the radiologists, i.e. visual comparison between MLAA images and CT, 100 PET/CT scans with minimal misalignment, i.e. minimal body motion between PET and CT, were collected. The data were split into 70 for training, 10 for validation, and 20 for evaluation. All the PET data were reconstructed using MLAA \cite{rezaei2016simultaneous} with 3 iterations and 21 subsets. Both $\mu$-MLAA and $\lambda$-MLAA were reconstructed with a voxel size of $2.036 \times 2.036 \times 2.027 mm^3$, followed by $5 mm$ in FWHM Gaussian post-smoothing. For image normalization, the $\lambda$-MLAA images were normalized by the mean value and then normalized using a hyperbolic tangent function, i.e. $\lambda_{norm} = tanh(\lambda / \hat{\lambda} / \sigma)$, where $\lambda$, $\hat{\lambda}$, and $\sigma$ are the $\lambda$-MLAA image, the mean value of the $\lambda$-MLAA image, and the hyper-parameter controlling the range of value, respectively. $\sigma$ was empirically set to 10 to ensure the organs of interest, i.e. except the bladder, are in a reasonable value zone for learning. On the other hand, $\mu$-MLAA and CT-derived $\mu$-map were normalized by $0.15$, which corresponds to the skull bone attenuation coefficient at 511 keV. On the other hand, we collected 2000 stand-alone CT scans, i.e. CT data excluding the patient used for training and evaluation in the previous part, for population data in PPGM. The data was also collected from YNHH. Subsequently, 2000 CT-derived $\mu$-map images were used and the same image normalization procedures were performed for this data. The network is implemented using Pytorch\footnote{http://pytorch.org/}. The Adam solver \cite{kingma2014adam} was used to optimize our network with the parameters $(\beta_1 , \beta_2) = (0.9, 0.999)$ and a learning rate of $1e-4$. A batch size of $6$ is used. For data augmentation, we randomly cropped 300 random patches for each patient data, resulting in 21000 training samples for patch-based training. 

\subsection{Evaluation Strategies and Baselines}
Using CT-derived $\mu$-map as the reference, we performed quantitative and qualitative evaluations on both 1) $\mu$-map generation and 2) AC PET based on the generated $\mu$-map. First, the quality of the generated $\mu$-map was evaluated using three image-quality metrics, including Peak Signal-to-Noise Ratio (PSNR), Structural Similarity Index (SSIM), and Root Mean Square Error (RMSE). SSIM focuses on the evaluation of structural recovery, while RMSE with a unit of Hounsfield unit (HU) and PSNR with a unit of dB stress the evaluation of intensity profile recovery. For comparative evaluation, we compared our results against previous DL-based AC methods, including UNet-based methods \cite{toyonaga2022deep,chen2021ct,ronneberger2015u} and GAN-based methods \cite{armanious2020independent,dong2019synthetic,isola2017image}. We also extended one of the current SOTA image restoration/generation methods to 3D, called MPRNet \cite{zamir2021multi} that also used multi-stage design, for comparison. For both quantitative and qualitative evaluations, we evaluated the performance under two different ultra-low-dose/count settings, i.e. 10\% and 2.5\% of full counts.  On the other hand, we further evaluated the $\mu$-map generation by applying them to PET reconstruction. Specifically, the $\mu$-map generated from low-count PET data, i.e. 10\% and 2.5\%, were used in full-count PET reconstruction with attenuation correction to reflect the quality of the $\mu$-map. Please note using ultra-low-count PET reconstruction for AC performance evaluation may be biased due to noisy reconstruction, thus full-count PET data was used here for PET AC performance evaluations. Here, PET reconstructions were performed using the OSEM algorithm (3 iterations and 21 subsets) implemented in the Siemens E7 toolkit.

%===========================================================
\section{Results}
\subsection{Experimental Results}
The comparative evaluation of our POUR-Net against prior DL-based PET AC methods across varying low-count scenarios is presented in Figure \ref{fig:comp_methods_umap}. In the first patient example at a 10\% low-count setting (1st row in Figure \ref{fig:comp_methods_umap}), the $\mu$-map generated by MLAA \cite{rezaei2016simultaneous} (i.e., $\mu$-MLAA) suffers from not only high noise level but artifacts. Using this $\mu$-MLAA and the corresponding $\lambda$-MLAA as input, the UNet-based method can generate a reasonable general profile of the $\mu$-map, but has difficulties in generating the fine detail structures, especially in the bone regions. On the other hand, OUR-Net using the exact same network input can produce $\mu$-map that contains much richer details, e.g. ribs, spine, and scapula (orange arrows in Figure \ref{fig:comp_methods_umap}). By further incorporating the OUR-Net with PPGM (i.e. POUR-Net), we can see the detail recovery is further improved, thus generating $\mu$-map with structure and intensity best matching with the CT-derived $\mu$-map on the last column. With the count/dose level further reduced to 2.5\% in the second patient example (2-rd row in Figure \ref{fig:comp_methods_umap}), we can see the $\mu$-map generated from MLAA suffers from enhanced noise and more severe artifacts, as compared to the 10\% low-count MLAA. Similarly, the UNet-based method, relying on $\mu$-MLAA and corresponding $\lambda$-MLAA as input, produces a reasonably clear $\mu$-map but with noticeably blurry structures owing to the challenging image quality of the input. In contrast, OUR-Net with the same input can produce $\mu$-map with much better quality. For instance, structures such as the bronchus near the heart (indicated by the orange arrow in the lung) missed by UNet are reasonably recovered by OUR-Net. Bone structures, including the spine and scapula, also exhibit better restoration with OUR-Net compared to the previous UNet-based method. The incorporation of PPGM into OUR-Net, forming POUR-Net, further refines detailed structures in the generated $\mu$-map under the challenging 2.5\% low-count scenario, aligning closely with the CT-derived $\mu$-map in the last column.

The quantitative assessment of different methods across two low-count settings is summarized in Table \ref{tab:comp_methods_umap}. In the 10\% low-count experiments, compared to the $\mu$-MLAA, we can observe that UNet-based method \cite{toyonaga2022deep} can significantly improve the $\mu$-map quality, elevating the PSNR from $26.82$dB to $33.75$dB for instance. Comparable performance enhancements are observed in contrast to other methodologies, including cGAN\cite{armanious2020independent}, RDUNet\cite{chen2021ct}, and MPRNet\cite{zamir2021multi}, all exceeding the baseline of $\mu$-MLAA. Notably, when compared to the leading performance from MPRNet, which also employs a multi-stage design, OUR-Net exhibits superior performance, achieving a PSNR increase from $33.92$ dB to $34.13$ dB. The integration of OUR-Net and PPGM in POUR-Net further amplifies the performance, yielding a remarkable PSNR of $37.01$dB, surpassing all prior benchmarks, including the standalone OUR-Net. Similar trends can be found in the 2.5\% low-count experiments, where $\mu$-MLAA quality degrades to PSNR $=22.51$dB due to count-starvation. Even in this challenging scenario, POUR-Net can generate high-quality $\mu$-maps with a PSNR of $35.12$dB, outperforming all compared baselines. Similar trends can be found when using other image evaluation metrics. 

\begin{figure}[htb!]
\centering
\includegraphics[width=0.48\textwidth]{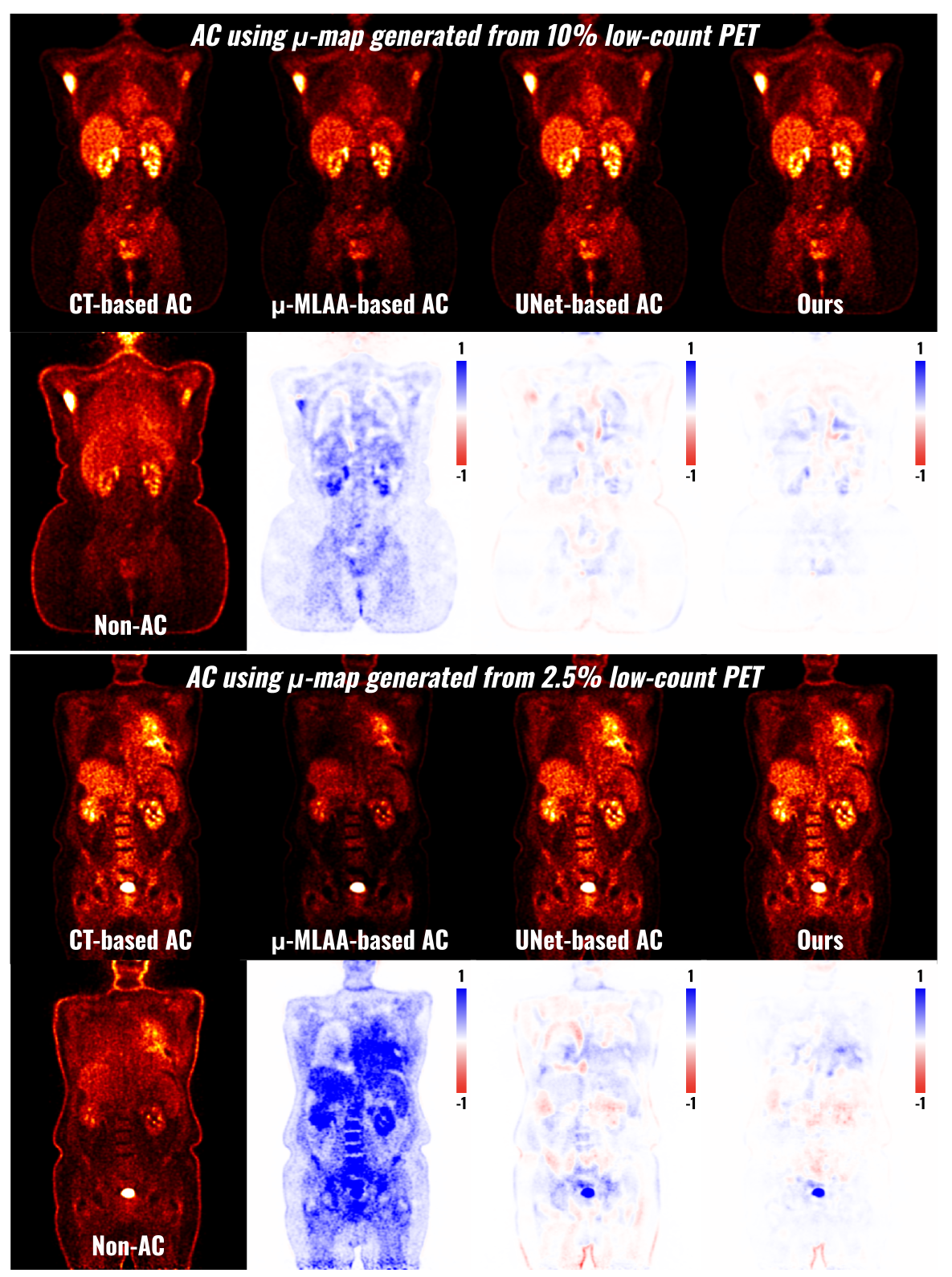}
\caption{Visual comparison of full-count PET reconstructions using $\mu$-map generated from low-count PET data. The non-attenuation-corrected PET reconstructions (bottom left) suffer from severe quantification errors, as compared to PET reconstructions with AC using CT-derived $\mu$-map (top left). Compared to prior $\mu$-map generation methods (central two columns), POUR-Net (last column) shows much better PET quantification accuracy.}
\label{fig:comp_methods_recon}
\end{figure}

\begin{table} [htb!]
\footnotesize
\centering
\caption{Quantitative comparison of full-count PET reconstructions using $\mu$-map generated from low-count PET data. Best results are marked in \textbf{bold}. "$\dagger$" means the difference between POUR-Net and all the baseline methods is significant at $p<0.001$.}
\label{tab:comp_methods_recon}
    \begin{tabular}{l|c|c|c}
        \hline
        \multirow{1}{*}{\textbf{Full-dose}}  & \multicolumn{3}{c}{\textbf{with $\mu$-map gen from 
 10\% low-count PET}}       \Tstrut\Bstrut\\
        \cline{2-4}
        \textbf{PET Recon}                   & PSNR            & SSIM              & RMSE         \Tstrut\Bstrut\\
        \hline
        $\mu$-MLAA \cite{rezaei2012simultaneous}                          & $36.88\pm2.86$   & $.981\pm.009$     & $.026\pm.012$    \Tstrut\Bstrut\\
        \hline
        UNet \cite{toyonaga2022deep}                                & $43.38\pm2.10$   & $.989\pm.004$     & $.006\pm.003$    \Tstrut\Bstrut\\
        \hline
        RDUNet \cite{chen2021ct}                               & $43.39\pm2.13$   & $.989\pm.005$     & $.006\pm.003$     \Tstrut\Bstrut\\
        \hline
        cGAN \cite{armanious2020independent}                                & $43.32\pm2.11$   & $.988\pm.006$     & $.007\pm.004$    \Tstrut\Bstrut\\
        \hline
        POUR-Net                                 & $\mathbf{44.88\pm.2.23}$$^\dagger$   & $\mathbf{.992\pm.003}$$^\dagger$    & $\mathbf{.004\pm.003}$$^\dagger$     \Tstrut\Bstrut\\
        \hline
        \hline
        \multirow{1}{*}{\textbf{Full-dose}}  & \multicolumn{3}{c}{\textbf{with $\mu$-map gen from 
 2.5\% low-count PET}}       \Tstrut\Bstrut\\
        \cline{2-4}
        \textbf{PET Recon}                   & PSNR            & SSIM              & RMSE         \Tstrut\Bstrut\\
        \hline
        $\mu$-MLAA \cite{rezaei2012simultaneous}                           & $30.17\pm2.97$   & $.965\pm.016$     & $.126\pm.062$    \Tstrut\Bstrut\\
        \hline
        UNet \cite{toyonaga2022deep}                                & $42.96\pm1.95$   & $.982\pm.004$     & $.008\pm.003$    \Tstrut\Bstrut\\
        \hline
        RDUNet \cite{chen2021ct}                              & $42.95\pm1.96$   & $.982\pm.004$     & $.008\pm.003$    \Tstrut\Bstrut\\
        \hline
        cGAN \cite{armanious2020independent}                                 & $42.89\pm1.93$   & $.980\pm.005$     & $.009\pm.003$    \Tstrut\Bstrut\\
        \hline
        POUR-Net                                 & $\mathbf{43.21\pm2.78}$$^\dagger$   & $\mathbf{.989\pm.004}$$^\dagger$    & $\mathbf{.005\pm.004}$$^\dagger$     \Tstrut\Bstrut\\
        \hline
    \end{tabular}
\end{table}

Moreover, we assessed PET reconstructed images with attenuation corrections using the $\mu$-maps generated as outlined in the previous paragraph. The patient examples of PET reconstruction, utilizing $\mu$-maps derived from 10\% and 2.5\% low-count PET data, are visually presented in Figure \ref{fig:comp_methods_recon}. As we can see, the non-attenuation-corrected (Non-AC) PET reconstruction (left bottom in the figure) suffers from severe quantification error, when compared to the CT-based AC result (left top in the figure) which is treated as the gold standard here. Although the use of $\mu$-MLAA for AC mitigates some attenuation effects, discernible biases persist in the resulting PET reconstruction, as indicated by the error maps. On the other hand, using $\mu$-map generated from previous methods, such as the UNet-based method, it shows a much-improved quantification accuracy in PET reconstruction. Notably, based on our POUR-Net, the quantification error is further reduced as can be clearly observed in the SUV error map.

The corresponding quantitative evaluation on PET reconstruction is summarized in Table \ref{tab:comp_methods_recon}. Aligning with the visual observations, the $\mu$-MLAA-based AC exhibits a notably elevated quantification bias error (e.g. RMSE of $0.026$ for 10\% low-count) when compared to previous DL-based methods (e.g. RMSE of $0.006$ from cGAN). The POUR-Net, producing the best quality $\mu$-maps among previous methods (Table \ref{tab:comp_methods_umap}) for AC, also provides the best PET reconstruction with significantly lower PET quantification errors among the previous baselines.

\begin{figure}[htb!]
\centering
\includegraphics[width=0.42\textwidth]{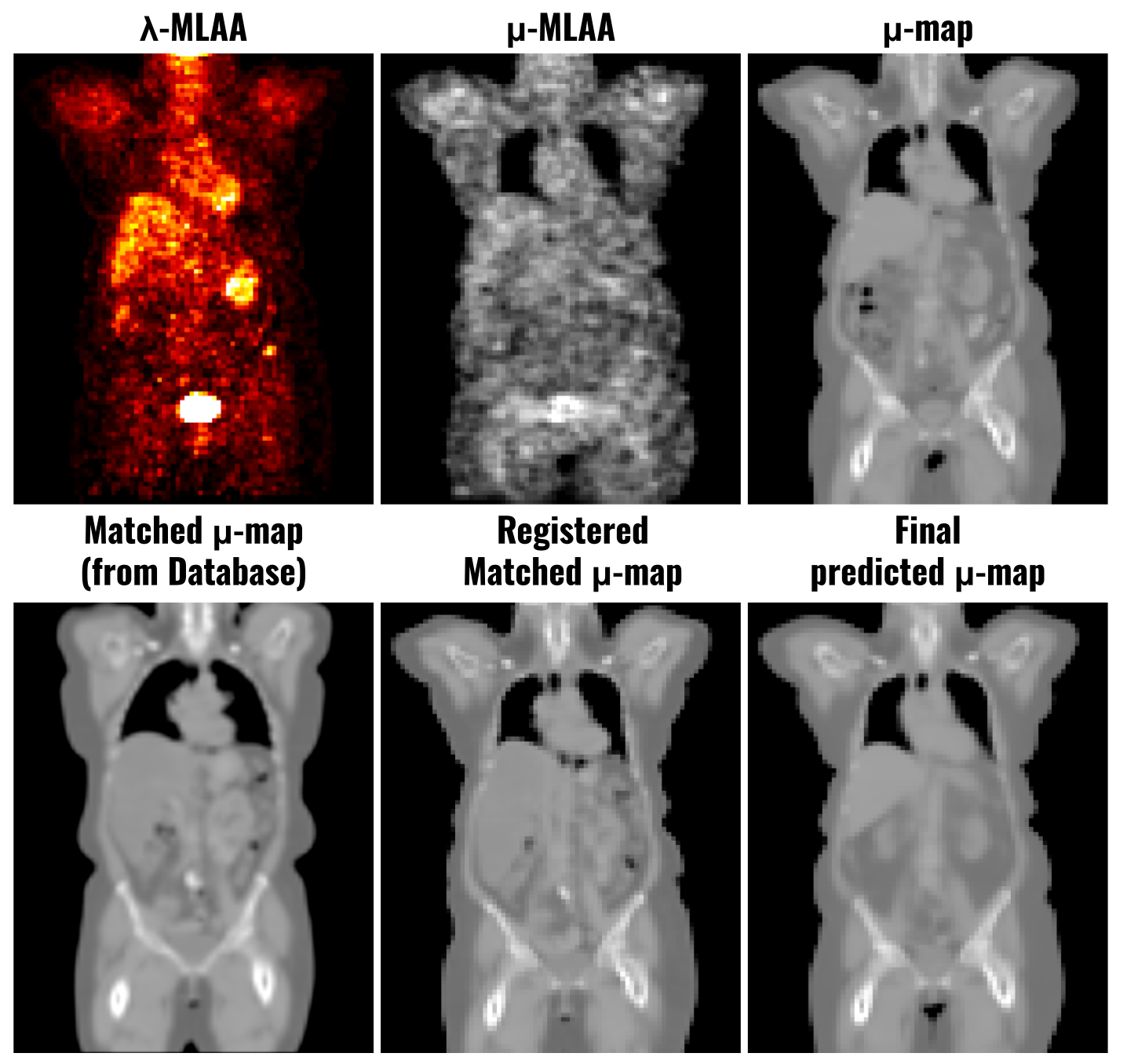}
\caption{Examples of input, intermediate output, and final outputs of the proposed PPGM. Using the OUR-Net prediction (bottom right), PPGM searches for the most match $\mu$-map (bottom left) and performs registration (bottom middle).}
\label{fig:PPGM}
\end{figure}

\subsection{Ablative Studies}
\textbf{1) Evaluation on Population Prior} The PPGM interleaved between OUR-Net significantly helped the $\mu$-map generation (Table \ref{tab:comp_methods_umap}). Here, we assessed the prior $\mu$-map generated from PPGM. Figure \ref{fig:PPGM} shows a 10\% low-count patient example. Given the MLAA reconstructions (top left \& middle) as input, PPGM searched for the most matched $\mu$-map from PPGM database (bottom left) and performed registration between it and the predicted $\mu$-map matched $\mu$-map (bottom right), generating the register matched $\mu$-map (bottom middle) for PPGM output. The corresponding quantitative evaluation is summarized in Table \ref{tab:PPGM}. Taking 10\% low-count as an example, we can see that the averaged PSNR between the most matched $\mu$-map and the CT-derived $\mu$-map (i.e. gold standard) is about $28.12$dB, indicating reasonable volumetric similarity. After registration, the PSNR is further boosted to $30.89$dB, and this brings additional info for $\mu$-map generation for the OUR-Net in the next cascade.  

\begin{table} [htb!]
\footnotesize
\centering
\caption{Quantitative evaluations of the intermediate and final outputs from PPGM under two different low-count PET settings. Please note the first PPGM from the cascade framework was assessed here.}
\label{tab:PPGM}
    \begin{tabular}{l|c|c|c}
        \hline
        \multirow{1}{*}{\textbf{Population}}  & \multicolumn{3}{c}{\textbf{10\% low-count PET}}       \Tstrut\Bstrut\\
        \cline{2-4}
        \textbf{Prior Evaluation}                   & PSNR            & SSIM              & RMSE            \Tstrut\Bstrut\\
        \hline
        Matched                               & $28.12\pm3.89$  & $.912\pm.033$     & $.035\pm.030$      \Tstrut\Bstrut\\
        \hline
        Registered Matched                    & $30.89\pm2.30$  & $0.957\pm.025$    & $.023\pm.012$   \Tstrut\Bstrut\\
        \hline
        \multirow{1}{*}{\textbf{Population}}  & \multicolumn{3}{c}{\textbf{2.5\% low-count PET}}      \Tstrut\Bstrut\\
        \cline{2-4}
        \textbf{Prior Evaluation}                   & PSNR            & SSIM              & RMSE            \Tstrut\Bstrut\\
        \hline
        Matched                               & $27.68\pm3.66$  & $.909\pm.036$     & $.039\pm.031$      \Tstrut\Bstrut\\
        \hline
        Registered Matched                    & $30.03\pm2.60$  & $0.945\pm.030$    & $.028\pm.019$   \Tstrut\Bstrut\\
        \hline
    \end{tabular}
\end{table}

\begin{figure}[htb!]
\centering
\includegraphics[width=0.38\textwidth]{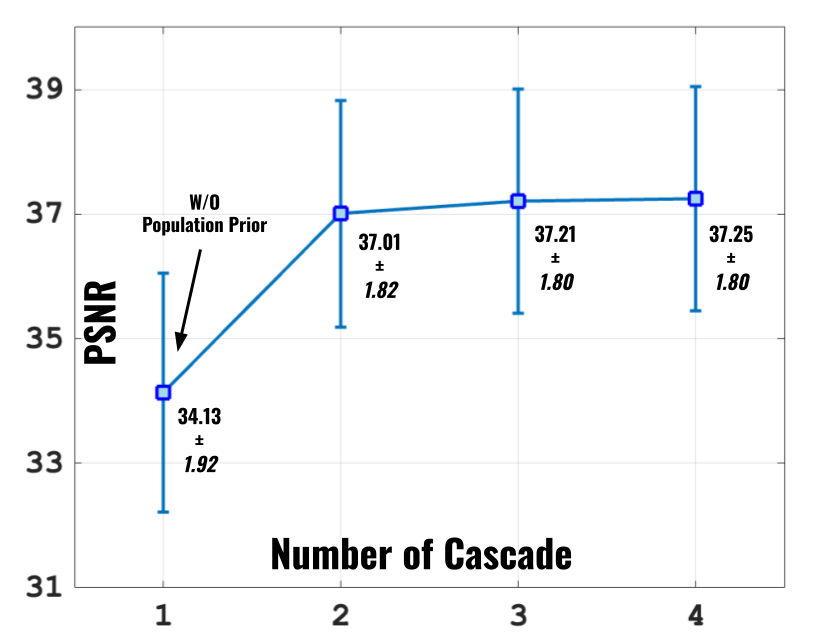}
\caption{Quantitative analysis of the impact of the number of cascades in POUR-Net. The number of cascades is annotated on the x-axis. 10\% low-count PET is considered here. }
\label{fig:plot_cascade}
\end{figure}

\textbf{2) Impact of Number of Cascade} POUR-Net uses a cascade design with multiple OUR-Net and PPGM interleaved. Here we further evaluated the performance when a different number of cascades were used in the POUR-Net. The performance is summarized in Figure \ref{fig:plot_cascade} based on 10\% low-count PET data. When the number of cascades equals 1, i.e. OUR-Net only without PPGM, the performance is about PSNR $=34.13$dB. As we increase the number of cascades to 2 with two OUR-Net and one PPGM interleaved, the performance gains a significant boost with PSNR increased to $37.01$dB. As we further increase the number of cascades in POUR-Net, the $\mu$-map generation performance starts to converge at around $37.21$dB. Similar trends were found when using other image quality evaluation metrics. 

\textbf{3) Impact of Representation Branches} OUR-Net is the main backbone network used in the POUR-Net framework, OUR-Net utilizes both under-representation and over-representation branches to aid the $\mu$-map generation. Thus, we further evaluated the impacts of the representation branches based on the 10\% low-count PET data. Specifically, we investigated the OUR-Net performance when different configurations of the representation branches were used. The analysis results are summarized in Table \ref{tab:branches}. When neither OvNet (i.e. O Branch) nor UnNet (i.e. U Branch) was included in OUR-Net, the OUR-Net with only FuNet that operates solely on the original resolution, achieves sub-optimal performance with PSNR $=33.66$. Adding OvNet or UnNet into OUR-Net both demonstrated improved performance of $\mu$-map generation. The best performance was found when both branches were included in the OUR-Net. Similar trends were found when using other image quality evaluation metrics. 

\begin{table} [htb!]
\footnotesize
\centering
\caption{Ablative studies on the inclusion of different representation learning in OUR-Net. \textcolor{green}{\cmark} and \xmark\space means over-represented branch (O-branch) or under-represented branch (U-branch) used and not used in OUR-Net, respectively. 10\% low-count PET is considered here.}
\label{tab:branches}
\begin{tabular}{|c c||c|c|c|}
    \hline
    O Branch                     & U Branch                    & PSNR                   & SSIM                   & RMSE               \Tstrut\Bstrut\\
    \hline   
    \xmark                       & \xmark                      & $33.66\pm2.01$         & $.969\pm.007$          & $.014\pm.010$    \Tstrut\Bstrut\\
    \textcolor{green}{\cmark}    & \xmark                      & $33.88\pm1.97$         & $.971\pm.006$          & $.013\pm.009$    \Tstrut\Bstrut\\
    \xmark                       & \textcolor{green}{\cmark}   & $34.01\pm1.95$         & $.977\pm.006$          & $.011\pm.008$    \Tstrut\Bstrut\\
    \textcolor{green}{\cmark}    & \textcolor{green}{\cmark}   & $34.13\pm1.92$         & $.979\pm.006$          & $.010\pm.008$    \Tstrut\Bstrut\\
    \hline
\end{tabular}
\end{table}

%===========================================================
\section{Discussion}
% Dicussion on our idea - mult-scale representation + population prior aiding generation

% Discussion on the results: 1) u-map generation 2) PET recon 3) Abation studies

% Discussion on limitations and future work: 
% 1) 3D training takes long, especially over-representation. and PPGM search could take time
% 2) Evaluation of clinically important findings, tumors with accurate contouring
% 3) Evaluation of other tracers

In this work, we developed a novel deep learning cascade framework, called POUR-Net, for low-count PET $\mu$-map generation, thus enabling accurate PET attenuation correction based on the generated $\mu$-map. Specifically, POUR-Net enables high-quality $\mu$-map generation with contributions from two main components, including (1) an over-under-representation network (OUR-Net) and (2) a population-prior generation machine (PPGM). First, OUR-Net is the backbone network used for $\mu$-map generation in each cascade step in our framework. OUR-Net can simultaneously learn abstracted feature representation (via the UnNet branch) and fine detail representation (via the OvNet branch), which are then combined to assist the $\mu$-map generation at the original resolution level (FuNet branch). As evidenced in Table \ref{tab:branches}, both under-representation and over-representation can help improve the $\mu$-map generation performance, and the best performance was achieved when both representation branches were used to assist, better than the previous baseline methods (Table \ref{tab:comp_methods_umap}). In fact, over-representation network designs have also shown promising performance in fine structure segmentation in prior works \cite{valanarasu2021kiu}. Second, we developed PPGM to further improve OUR-Net generation by bringing additional $\mu$-map information from a stand-alone and large-scale population CT-derived $\mu$-map dataset. Using an initial prediction from OUR-Net as the reference, PPGM retrieves the most match $\mu$-map from the dataset, and then registers it to the initial prediction (Table \ref{tab:PPGM}). The registered $\mu$-map along with the original low-count PET data is inputted again into OUR-Net for $\mu$-map generation in the next cascade, thus significantly improving the $\mu$-map generation performance (Figure \ref{fig:plot_cascade}). Overall, we found that POUR-Net benefits more from PPGM with approximately a 2dB PSNR performance boost. From Table \ref{tab:comp_methods_umap}, we can see using POUR-Net with the proposed cascaded design generated significantly better $\mu$-map quality than previous DL-based methods, in terms of image quality metrics. Furthermore, using the generated $\mu$-map for PET reconstruction with AC, we can see the PET quantification error based on POUR-Net generated $\mu$-map was also significantly lower than previous DL-based methods (Table \ref{tab:comp_methods_recon}). 

Our current work also has potential limitations, suggesting interesting directions for future studies. First, the training and inference computation cost was relatively high compared to previous DL-based methods. The OUR-Net used in our framework is a 3D network with an over-representation branch. The over-represented feature maps in OvNet require longer computation time for convolution operations, given the large volume size after upsampling (i.e. $\times$4 the input size at encoder output). Therefore, the training of OUR-Net is approximately 5 times longer than UNet-based methods, and took us about 100 hours to complete the training. On the other hand, the searching process in PPGM introduced additional inference time. In the current implementation, we iterated through all the patient volumes in PPGM to identify the one with minimal NMSE, taking about 10 mins using one PPGM. The time increases linearly as the number of cascades in POUR-Net increases. Using a POUR-Net with two cascades, the average time for generating each $\mu$-map took about 15 mins which is in a reasonable range. In the future, we will investigate strategies to accelerate training and inference. Second, the evaluations throughout this work focused on the overall assessment of image quality, using image quality metrics such as PSNR, SSIM, and NMSE. Evaluations on clinically important findings, e.g. tumors and inflammation, on both $\mu$-map and PET reconstruction were not included here, and will be extensively investigated in our future works. Finally, our work mainly focused on network development and evaluations on \textsuperscript{18}F-FDG PET tracer, and did not include other tracer types. \textsuperscript{18}F-FDG is the most widely used tracer in PET, and we believe our method can be adapted to other tracer types given enough data for training. Please also note that PPGM is independent of tracer type, thus allowing adaptation to any tracer type. Evaluation of other PET tracers will be an important future direction. 

\section{Conclusion}
In this paper, we introduce POUR-Net, a population-prior-aided over-under-representation network for generating high-quality attenuation maps ($\mu$-map) from low-count/dose PET data. Comprising OUR-Net as the backbone and assisted by PPGM in the cascade framework, POUR-Net demonstrated superior $\mu$-map generation quality across different low-count scenarios, as well as better PET quantification against previous baseline methods from our experiments. We believe that our proposed method can potentially enable CT-free low-count PET attenuation correction, thus further reducing the radiation dose in low-dose PET/CT imaging. 

%===========================================================
% \section*{Acknowledgment}
% This work was supported by funding from the National Institutes of Health (NIH) under grant numbers R01EB025468, R01CA224140, and R01CA275188. 

%===========================================================
% \clearpage
\bibliographystyle{IEEEtran}
\bibliography{bibliography}

\end{document}